\documentclass[conference]{IEEEtran}
\IEEEoverridecommandlockouts
\usepackage{cite}
\usepackage{amsmath,amssymb,amsfonts}
\usepackage{algorithmic}
\usepackage{graphicx}
\usepackage{textcomp}
\usepackage{xcolor}
\usepackage{booktabs}
\usepackage{multirow} 
\usepackage{subcaption}
\def\BibTeX{{\rm B\kern-.05em{\sc i\kern-.025em b}\kern-.08em
    T\kern-.1667em\lower.7ex\hbox{E}\kern-.125emX}}

\begin{document}

\title{TokenSeg: Efficient 3D Medical Image Segmentation via Hierarchical Visual Token Compression\\
{}
\thanks{}
}

\author{
Sen Zeng$^1$, Hong Zhou$^2$, Zheng Zhu$^3$, Yang Liu$^4$ \\
$^1$Tsinghua University \quad $^2$Southwest Forestry University \quad $^3$GigaAI\quad $^4$KCL\\[0.3em]
\texttt{zengsen2024@gmail.com, 5515@swfu.edu.cn, zhengzhu@ieee.org, yang.9.liu@kcl.ac.uk} \\

}

\maketitle

\begin{abstract}
Three-dimensional medical image segmentation is a fundamental yet computationally demanding task due to the cubic growth of voxel processing and the redundant computation on homogeneous regions. To address these limitations, we propose \textbf{TokenSeg}, a boundary-aware sparse token representation framework for efficient 3D medical volume segmentation. Specifically, (1) we design a \emph{multi-scale hierarchical encoder} that extracts 400 candidate tokens across four resolution levels to capture both global anatomical context and fine boundary details; (2) we introduce a \emph{boundary-aware tokenizer} that combines VQ-VAE quantization with importance scoring to select 100 salient tokens, over 60\% of which lie near tumor boundaries; and (3) we develop a \emph{sparse-to-dense decoder} that reconstructs full-resolution masks through token reprojection, progressive upsampling, and skip connections. Extensive experiments on a 3D breast DCE-MRI dataset comprising 960 cases demonstrate that TokenSeg achieves state-of-the-art performance with 94.49\% Dice and 89.61\% IoU, while reducing GPU memory and inference latency by 64\% and 68\%, respectively. To verify the generalization capability, our evaluations on MSD cardiac and brain MRI benchmark datasets demonstrate that TokenSeg consistently delivers optimal performance across heterogeneous anatomical structures. These results highlight the effectiveness of anatomically informed sparse representation for accurate and efficient 3D medical image segmentation.
\end{abstract}

\begin{IEEEkeywords}
3D medical image segmentation, Sparse token representation, Boundary-aware segmentation, Computational efficiency
\end{IEEEkeywords}

\section{Introduction}
\label{sec:intro}

\begin{figure}[t]
	\centering
	\includegraphics[width=\columnwidth]{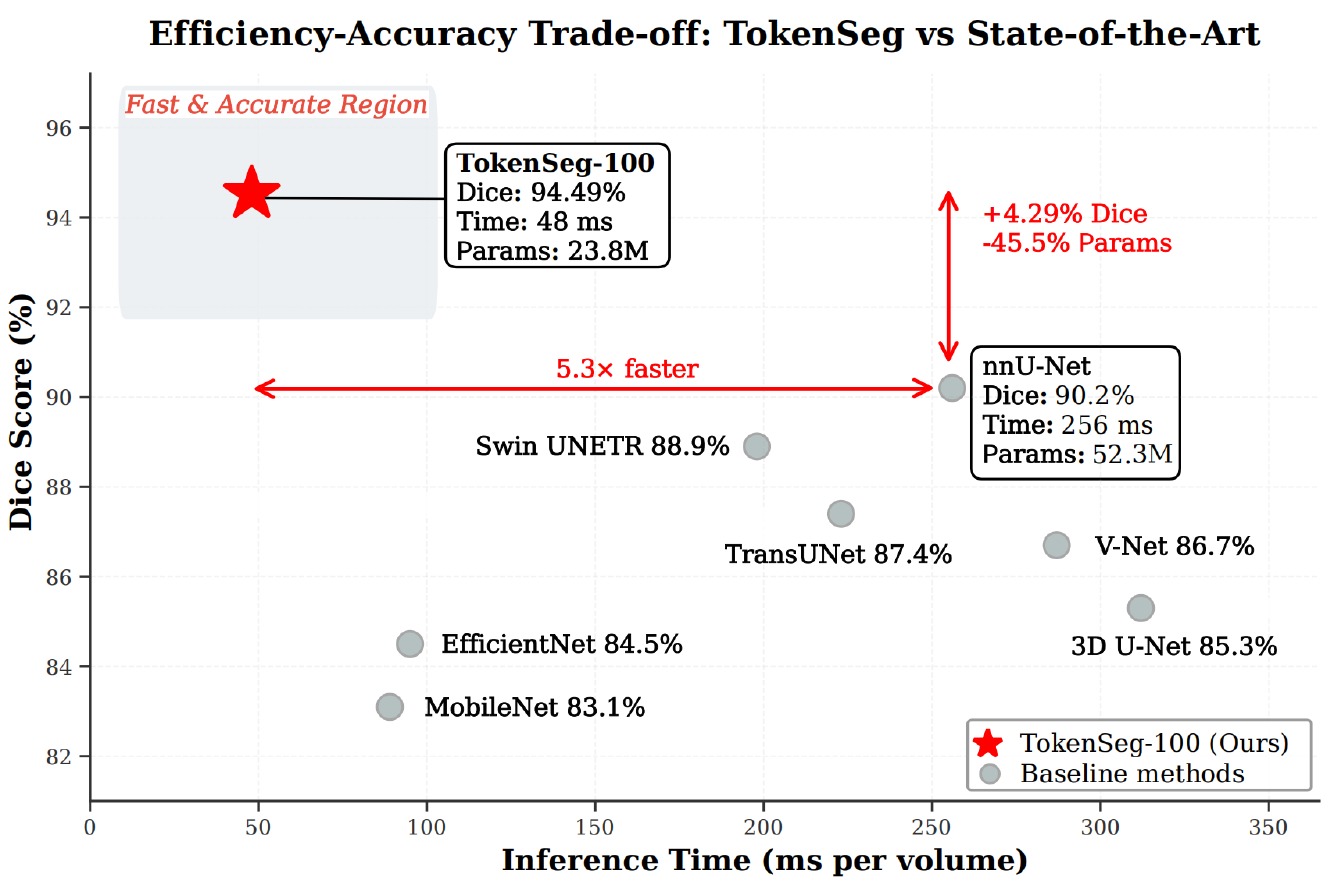}
	\caption{Performance comparison between TokenSeg and state-of-the-art methods on 3D medical volume segmentation. TokenSeg-100 achieves 94.49\% Dice score, outperforming the best baseline (nnU-Net) by +4.29\% Dice while being 5.3× faster (48ms vs 256ms) with 54.5\% fewer parameters (23.8M vs 52.3M).}
	\label{fig1}
\end{figure}

Three-dimensional medical image segmentation plays a pivotal role in modern clinical workflows, enabling precise delineation of anatomical structures and pathological lesions for diagnosis, treatment planning, and surgical navigation~\cite{b18,b19,b20}. Among various applications, automated tumor segmentation from volumetric modalities such as CT, MRI, and PET is crucial for quantitative assessment and personalized treatment. Despite the success of deep learning, achieving accurate and efficient 3D segmentation remains challenging due to the inherently cubic growth of computational complexity with respect to volume size.

Early deep learning-based methods such as U-Net~\cite{b16}, 3D U-Net~\cite{b8}, V-Net~\cite{b9}, nnU-Net~\cite{b10}, and derivatives like UNet++~\cite{b2} and FCN~\cite{b17} achieve strong performance by densely processing all voxels at full spatial resolution. However, these dense prediction paradigms are computationally inefficient and memory-intensive, as most voxels belong to homogeneous regions (e.g., air or fat) that contribute little to segmentation accuracy. Uniform processing not only wastes computation but also restricts scalability to high-resolution volumes, forcing patch-based or downsampled inference that compromises global context and boundary precision, two factors essential in clinical practice~\cite{b33}.

To address these issues, recent research has explored efficient or sparse modeling strategies. Sparse convolutional networks~\cite{b21} and conditional computation/dynamic routing~\cite{b23,b24,b22} selectively activate computation in salient regions, while attention- and anatomy-guided networks emphasize organ- or lesion-specific features~\cite{b1,b25,b26,b27,b28}. Although these approaches reduce redundancy, they primarily operate at the feature or voxel level and lack an explicit compact representation of volumetric data. Meanwhile, the emergence of vision transformers (ViTs)~\cite{b7,b29} has introduced token-based modeling to medical imaging~\cite{b13,b30,b31,b12}, enabling long-range dependency learning. Yet, their dense tokenization (e.g., $16^3$ patches guided by hierarchical backbones such as Swin~\cite{b32}) remains computationally heavy and overlooks boundary-aware prioritization, which is critical for delineating lesion margins.

In the broader vision community, visual token compression has shown that dense pixel-level processing is not always necessary. DeepSeek-OCR~\cite{b11}, for example, compresses 4K document images into a few hundred tokens via vector quantization and importance scoring, achieving comparable recognition accuracy with a fraction of the computation. Inspired by this paradigm, we explore whether a similar compression principle can be extended to volumetric medical data. However, unlike 2D document understanding, 3D medical segmentation requires spatially coherent predictions and precise boundary delineation; diagnostic reliability depends on accurately capturing lesion margins rather than global texture cues~\cite{b34,b35,b36,b37}. This motivates a framework that not only compresses volumetric data effectively but also preserves anatomically meaningful structures during token selection and reconstruction.

To this end, we propose \textbf{TokenSeg}, a boundary-aware sparse token representation framework for efficient 3D medical image segmentation. TokenSeg introduces three key designs: (1) a \emph{multi-scale hierarchical encoder} that extracts candidate tokens across four resolution levels to capture both global anatomical context and fine-grained boundary details; (2) a \emph{boundary-aware tokenizer} that combines vector-quantized representation~\cite{b38} with importance-based selection to retain only the most informative tokens concentrated around anatomical boundaries~\cite{b39}; and (3) a \emph{sparse-to-dense decoder} that reconstructs high-resolution segmentation masks through token reprojection, progressive upsampling, and skip connections. As shown in Fig.~\ref{fig1}, extensive experiments on a large-scale breast DCE-MRI dataset demonstrate that TokenSeg achieves state-of-the-art segmentation accuracy while reducing GPU memory and inference latency by over 60\%. The results validate that anatomically informed sparse representation enables efficient and accurate 3D medical image segmentation.

\section{Method}
\label{sec:formatting}

\begin{figure*}[t!]
	\centering
	\includegraphics[width=\textwidth]{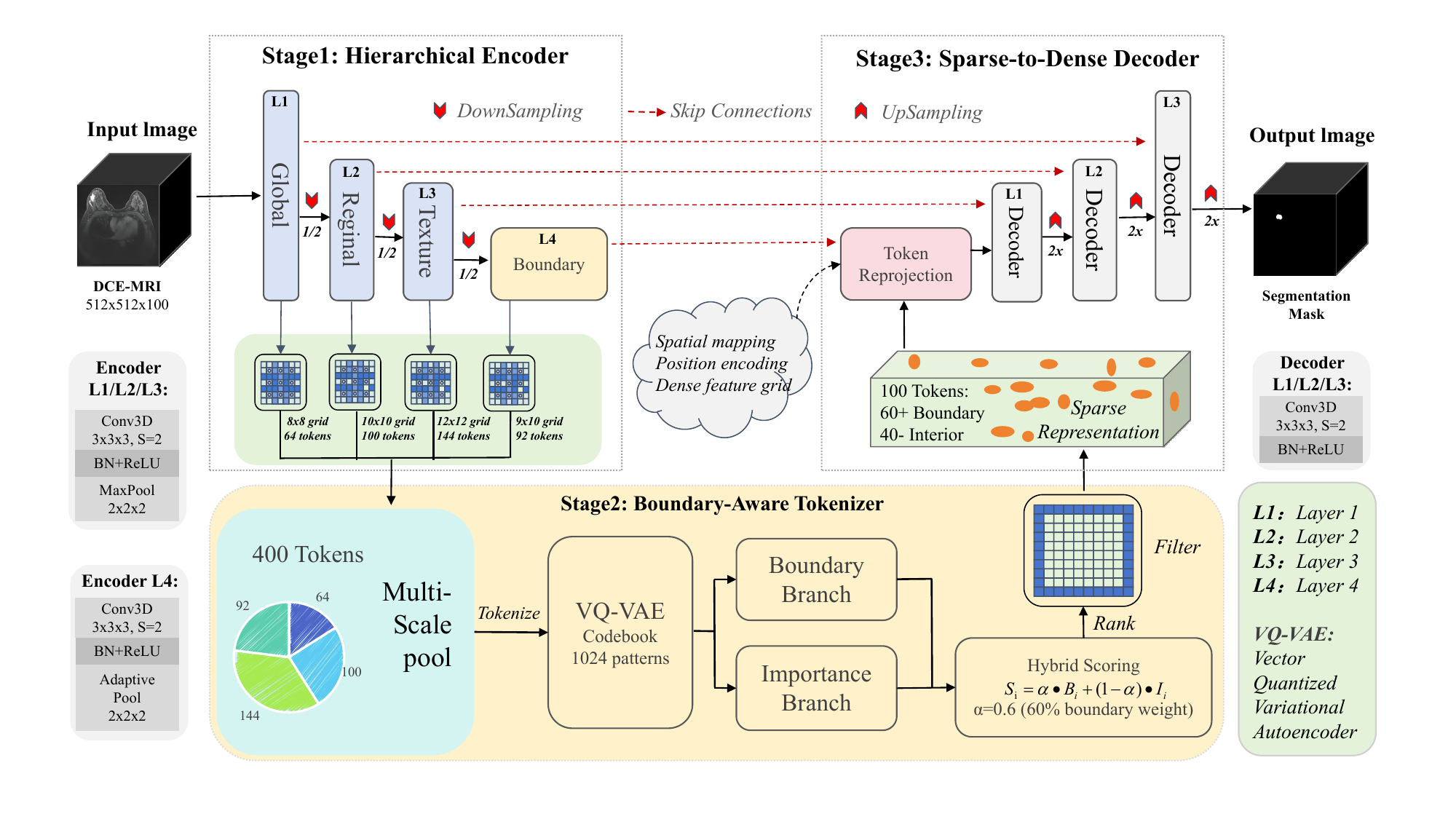} %
	\caption{The architecture of TokenSeg for the DCE-MRI Breast Cancer segmentation} 
\label{fig:main_framework}
\end{figure*}

Figure~\ref{fig:main_framework} illustrates our proposed \textbf{TokenSeg} architecture. The model takes a volumetric input $\mathbf{X}\in\mathbb{R}^{D\times H\times W}$ (single–channel DCE–MRI, typically $512\times512\times100$) and predicts a binary segmentation $\hat{\mathbf{Y}}\in\{0,1\}^{D\times H\times W}$. Unlike dense 3D CNN/ViT pipelines that uniformly aggregate local and global features over all voxels, thereby incurring prohibitive memory and latency, TokenSeg replaces heavy, full-field aggregation with a compact, boundary-centric token flow. Concretely, a \emph{hierarchical encoder} first converts $\mathbf{X}$ into a small multi-scale pool of candidate tokens that capture global-to-local cues; a \emph{boundary-aware tokenizer} then records only the task-critical tokens near anatomical margins via vector-quantized prototyping and a lightweight importance ranking; finally, a \emph{sparse-to-dense decoder} reprojects the selected tokens back to their spatial anchors and progressively reconstructs a full-resolution mask with skip-assisted refinement. This design targets the core bottleneck of volumetric segmentation, computing heavily where information concentrates (boundaries) while avoiding redundant processing on homogeneous tissue, achieving extreme spatial compression without sacrificing margin precision. We describe the three components in the following sections.


\subsection{Hierarchical Encoder}

Accurate volumetric segmentation requires \emph{global context} to constrain plausible shapes and \emph{local evidence} to resolve margins; operating at a single scale either loses detail (coarse) or becomes prohibitively expensive (fine). 
We therefore build a multi-level feature pyramid with \(L=4\) resolutions indexed by \(\ell\in\{1,\dots,L\}\) and spatial factors \(2^{-\ell}\), so that deeper levels summarize organ-level context while shallower levels preserve boundary cues. 
To convert dense features into a bounded sequence amenable to selection, each level is partitioned into non-overlapping local cells over the spatial lattice and each cell is pooled into a token; concatenating across the \(L\) levels yields a compact multi-scale \emph{candidate pool} with \(N=400\) tokens in total. 
This transforms the volume into a semantics-preserving representation that retains boundary evidence while avoiding the cubic cost of uniformly processing all voxels.


\subsection{Boundary-Aware Tokenizer}
\label{sec:tokenizer}

Volumetric MRI is dominated by background and large homogeneous regions with limited discriminative value, whereas segmentation accuracy is decided at \emph{label transitions} (anatomical boundaries). Allocating equal budget to all \(N\) candidate tokens thus disperses computation to blank or low–contrast areas and weakens boundary modeling. We therefore adopt a boundary–prioritized tokenizer that concentrates capacity near margins while keeping the representation stable across scans.

We are inspired by the visual token–compression paradigm of DeepSeek-OCR~\cite{b11}, which reduces dense processing through vector quantization and importance ranking. Unlike document recognition, where loose spatial correspondence is acceptable and legibility is the target, medical segmentation requires spatially coherent masks and precise margins. Accordingly, our tokenizer (i) \emph{stabilizes} token representations before ranking and (ii) \emph{biases} selection toward boundary-adjacent evidence; spatial anchoring is then preserved by the decoder.

Let \(\mathcal{T}_{\text{pool}}=\{\mathbf{t}_i\}_{i=1}^{N}\) denote the multi-scale candidate tokens emitted by the encoder; each token \(\mathbf{t}_i\in\mathbb{R}^{C}\) is a \(C\)-dimensional feature vector associated with a spatial location on some pyramid level. Our goal is to select a sparse subset \(\mathcal{T}_{\text{sparse}}\subset\mathcal{T}_{\text{pool}}\) of size \(|\mathcal{T}_{\text{sparse}}|=K\) with \(K\ll N\), retaining boundary-critical evidence while suppressing redundancy.

To make ranking comparable across volumes and robust to acquisition variability, we discretize tokens using a learnable codebook of visual prototypes. Let
\(\mathcal{C}=\{\mathbf{c}_k\in\mathbb{R}^{C}\}_{k=1}^{M}\) be the codebook with \(M\) prototypes. Each token \(\mathbf{t}_i\) is assigned to its nearest prototype in Euclidean distance,
\[
\mathbf{t}_i^{q} \;=\; \mathbf{c}_{k^\star(i)}, 
\qquad 
k^\star(i) \;=\; \arg\min_{k\in\{1,\dots,M\}} \big\|\mathbf{t}_i-\mathbf{c}_k\big\|_2,
\]
and the encoder–codebook pair is trained with the standard vector-quantization objective~\cite{b38} (embedding loss plus a commitment term). This discretization (a) collapses scanner/protocol idiosyncrasies into shared prototypes, (b) suppresses spurious activations due to noise or motion, and (c) yields discrete identities whose usage counts
\(\mathrm{freq}(\mathbf{t}_i^{q})\in\mathbb{N}\) (the number of candidates mapped to the same prototype) can be tracked to control redundancy.

Given the quantized token \(\mathbf{t}_i^{q}\), we score each candidate by jointly favoring semantic strength, boundary proximity, and prototype diversity:
\[
\mathrm{Score}(\mathbf{t}_i)
\;=\;
\|\mathbf{t}_i^{q}\|_2
\;\cdot\;
P_{\mathrm b}(\mathbf{t}_i)
\;\cdot\;
\log\!\Big(\tfrac{N}{\mathrm{freq}(\mathbf{t}_i^{q})}\Big).
\]
Here \(\|\mathbf{t}_i^{q}\|_2\in\mathbb{R}_{\ge 0}\) measures token strength; 
\(P_{\mathrm b}(\mathbf{t}_i)\in[0,1]\) is a scale-normalized boundary-proximity estimate computed on the token's pyramid level from local edge/gradient evidence around its spatial origin~\cite{b39}; and \(\mathrm{freq}(\mathbf{t}_i^{q})\) penalizes ubiquitous prototypes in an IDF-style manner. 
We then define the sparse set by top-\(K\) selection:
\[
\mathcal{T}_{\text{sparse}}
\;=\;
\operatorname*{TopK}_{\mathbf{t}_i\in\mathcal{T}_{\text{pool}}}
\mathrm{Score}(\mathbf{t}_i),
\qquad K\ll N.
\]

The proposed tokenizer concentrates computation where supervision and uncertainty peak (near boundaries), while prototype discretization~\cite{b5,b6,b38} makes scores stable across acquisitions. In practice, we use \(N=400\) candidates and retain \(K=100\), providing boundary-critical evidence for the following spatially anchored reconstruction.


\subsection{Sparse-to-Dense Decoder}
\label{sec:decoder}

Reconstructing a spatially coherent mask from a sparse set of \(K\) tokens requires maintaining global anatomical plausibility while turning discrete evidence near boundaries into continuous surfaces. Our decoder achieves this by (i) reprojecting tokens to their native pyramid lattices as \emph{spatial anchors}, (ii) progressively restoring resolution with cross-level fusion to propagate context and detail, and (iii) producing a calibrated dense probability volume at full resolution.

\paragraph{Token reprojection as spatial anchors.}
Let \(\mathcal{T}_{\text{sparse}}=\{\mathbf{t}_i\}_{i=1}^{K}\) be the selected tokens (Sec.~\ref{sec:tokenizer}). 
Each token \(\mathbf{t}_i\in\mathbb{R}^{C_{s_i}}\) is tied to a pyramid level \(s_i\in\{1,\dots,L\}\) (downsampling factor \(2^{-s_i}\)) and a lattice coordinate \((d_i,h_i,w_i)\) on that level. 
For each level \(s\), we initialize a sparse feature grid \(\widetilde{\mathbf{F}}^{(s)}\in\mathbb{R}^{D_s\times H_s\times W_s\times C_s}\) by placing tokens back at their original coordinates and setting all other sites to zero:
\[
\widetilde{\mathbf{F}}^{(s)}(d,h,w) \;=\;
\begin{cases}
\mathbf{t}_i, & \text{if } s=s_i \text{ and } (d,h,w)=(d_i,h_i,w_i),\\
\mathbf{0}, & \text{otherwise}.
\end{cases}
\]
Here \((D_s,H_s,W_s)=(\lfloor 2^{-s}D\rfloor,\lfloor 2^{-s}H\rfloor,\lfloor 2^{-s}W\rfloor)\) and \(C_s\) is the channel dimension at level \(s\). 
These anchors preserve topology and provide a scaffold for interpolation rather than hallucinating shapes from a bag of points.

\paragraph{Progressive reconstruction with cross-level fusion.}
Starting from the coarsest level, the decoder upsamples features by a factor of two per stage while fusing the corresponding encoder features to inject semantics and boundary detail. 
Let \(\mathbf{G}^{(L)}=\phi\!\big(\widetilde{\mathbf{F}}^{(L)}\big)\) be the decoded feature at the coarsest level after a local refinement operator \(\phi(\cdot)\) (a small stack of \(3{\times}3{\times}3\) convolutions). 
For stages \(s \in \{L-1,\dots,1\}\), we compute
\[
\mathbf{G}^{(s)} \;=\; \psi\!\Big(
\,\mathrm{Concat}\big(\,\mathcal{U}_2(\mathbf{G}^{(s+1)}),\, \mathbf{F}_{\text{enc}}^{(s)}\,\big)\,
\Big),
\]
where \(\mathcal{U}_2(\cdot)\) is a \(2\times\) upsampling operator defined on 3D lattices (e.g., trilinear or learned), \(\mathbf{F}_{\text{enc}}^{(s)}\) denotes the encoder feature at level \(s\) (skip connection), \(\mathrm{Concat}(\cdot,\cdot)\) concatenates along channels, and \(\psi(\cdot)\) is a refinement operator analogous to \(\phi(\cdot)\). 
This scheme (i) lifts coarse global context upward, (ii) injects high-frequency cues around boundaries via skips, and (iii) fills non-anchored regions smoothly under multi-scale guidance.

\paragraph{Dense mask prediction.}
At full resolution (\(s=1\)), a pointwise prediction head converts \(\mathbf{G}^{(1)}\) into a calibrated probability volume:
\[
\widehat{\mathbf{Y}} \;=\; \sigma\!\big(\,\Theta(\mathbf{G}^{(1)})\,\big)\in[0,1]^{D\times H\times W},
\]
where \(\Theta(\cdot)\) is a \(1{\times}1{\times}1\) linear projection and \(\sigma(\cdot)\) is the sigmoid function. 
A threshold of \(\theta\) is used at inference to obtain the binary prediction \(\widehat{\mathbf{Y}}\).

\paragraph{Properties.}
Reprojection preserves spatial correspondence and avoids topological shortcuts; progressive upsampling with skips counteracts the token bottleneck by reintroducing boundary detail at each scale; the final pointwise head produces a well-calibrated dense mask. 
Together, these choices translate a sparse, boundary-centric representation into a coherent segmentation while respecting both global anatomy and fine margins.

\subsubsection{Loss Function}

We optimize TokenSeg with a compact objective that couples overlap, calibration, and prototype stability:
\[
\label{eq:total-loss}
\mathcal{L}_{\text{total}}
=
\lambda_{\text{dice}}\mathcal{L}_{\text{Dice}}
+
\lambda_{\text{bce}}\mathcal{L}_{\text{BCE}}
+
\lambda_{\text{vq}}\mathcal{L}_{\text{VQ}}.
\]

\noindent\textbf{Overlap term.}
Let \(\Omega\) be the voxel index set, \(y_i\in\{0,1\}\) the ground-truth label, and \(\hat{y}_i\in[0,1]\) the predicted probability at voxel \(i\in\Omega\). The soft Dice loss directly optimizes the evaluation metric while being robust to foreground imbalance:
\[
\mathcal{L}_{\text{Dice}}
=
1-
\frac{2\sum_{i\in\Omega} y_i \hat{y}_i + \epsilon}
{\sum_{i\in\Omega} y_i + \sum_{i\in\Omega} \hat{y}_i + \epsilon},
\]
with a small \(\epsilon>0\) for numerical stability.

\noindent\textbf{Calibration term.}
To provide fine-grained voxel-wise supervision and improve probability calibration, we add a binary cross-entropy term
\[
\label{eq:bce}
\mathcal{L}_{\text{BCE}}
=
-\frac{1}{|\Omega|}
\sum_{i\in\Omega}
\big[
y_i\log \hat{y}_i + (1-y_i)\log(1-\hat{y}_i)
\big].
\]

\noindent\textbf{Prototype stability term.}
Denote by \(\mathcal{T}_{\text{pool}}=\{\mathbf{t}_j\}_{j=1}^{N}\) the candidate tokens and by \(\mathbf{t}_j^{q}\) their vector-quantized prototypes from the codebook \(\mathcal{C}\). The vector-quantization objective~\cite{b38} jointly trains the encoder and the codebook to yield stable prototypes:
\[
\label{eq:vq}
\mathcal{L}_{\text{VQ}}
=
\frac{1}{N}
\sum_{j=1}^{N}
\Big(
\underbrace{\|\mathbf{t}_j - \mathrm{sg}[\mathbf{t}_j^{q}]\|_2^{2}}_{\text{codebook (embedding) loss}}
+
\beta\underbrace{\|\mathrm{sg}[\mathbf{t}_j]-\mathbf{t}_j^{q}\|_2^{2}}_{\text{commitment loss}}
\Big),
\]
where \(\mathrm{sg}[\cdot]\) is the stop-gradient operator and \(\beta>0\) balances codebook usage and encoder commitment. This term reduces acquisition-induced variability, suppresses spurious activations, and enables reliable frequency-based diversity in the tokenizer.


\section{Experiments}

\subsection{Dataset Construction}

\noindent\textbf{Dataset Overview.}
We employed a large-scale private multi-center breast DCE-MRI dataset comprising 960 cases sourced from multiple institutions, partitioned into internal data (872 cases) and external data (88 cases from different centers). The internal data follows a 70\%-10\%-20\% split protocol, while the external dataset serves as a multi-center test set for cross-institutional generalization assessment. Additionally, we conducted evaluations on public benchmark datasets from the Medical Segmentation Decathlon (MSD)~\cite{b33}, including Task01 (brain glioma segmentation with 484 T1-weighted MRI scans) and Task02 (left atrium segmentation with 20 cardiac MRI scans). All volumes were preprocessed into uniform single-channel 3D data with consistent spatial resolution and intensity normalization. The total number of MRI scans utilized for training and testing in our study is 1,464. Detailed descriptions of these datasets and the preprocessing pipeline are provided in the supplementary material.

\subsection{Experimental Setup}

\paragraph{Implementation details.}
 
We implement TokenSeg in PyTorch~2.5.1(Python~3.12) and conduct all experiments on a single NVIDIA A800 GPU (80\,GB VRAM) with an Intel Xeon Platinum 8358P 8-core CPU and 256\,GB RAM. Optimization employs AdamW with a cosine-annealing learning rate scheduler. The initial learning rate is set to $10^{-4}$ and decays to $10^{-6}$; AdamW betas are $(\beta_1, \beta_2) = (0.9, 0.999)$ with weight decay $10^{-5}$. The per-GPU batch size is 2. Training runs for a maximum of 300 epochs with early stopping (patience $= 30$). We employ mixed precision training (FP16) via automatic mixed precision (AMP) to enhance computational efficiency, achieving approximately 40\% speedup. Data loading is optimized with 8 parallel workers and pinned memory for efficient GPU transfer.
Loss function coefficients are set to $\lambda_{\text{dice}} = 1.0$, $\lambda_{\text{bce}} = 0.5$, and $\lambda_{\text{vq}} = 0.1$. The vector-quantization commitment weight is $\beta = 0.25$, and the numerical stabilizer is $\epsilon = 10^{-5}$. The hierarchical encoder employs $L = 4$ pyramid levels. Multi-scale tokenization generates $N = 400$ candidate tokens per volume, from which the boundary-aware tokenizer selects the top $K = 100$ tokens to form the sparse representation for decoding. Threshold $\theta = 0.5$ is used at inference.

\subsection{Evaluation Metrics}

TokenSeg is evaluated across three dimensions: segmentation accuracy, computational efficiency, and compression quality, as detailed in Table~\ref{tab:metrics}. 

\begin{table}[h]
\centering
\caption{Comprehensive evaluation metrics.}
\label{tab:metrics}
\footnotesize
\setlength{\tabcolsep}{3pt}
\begin{tabular}{@{}llll@{}}
\toprule
\textbf{Metric} & \textbf{Target} & \textbf{Unit} & \textbf{Description} \\
\midrule
\multicolumn{4}{@{}l@{}}{\textit{Segmentation Performance}} \\
DSC & $>92\%$ & \% & Primary overlap metric \\
HD95 & $<5$ & mm & Boundary precision \\
Sensitivity & $>94\%$ & \% & Lesion detection rate \\
Precision & $>90\%$ & \% & False positive control \\
\addlinespace[0.3ex]
\multicolumn{4}{@{}l@{}}{\textit{Computational Efficiency}} \\
Inference Time & $<50$ & ms & Per-volume latency \\
GPU Memory & $<3$ & GB & Peak memory usage \\
Parameters & $<100$ & M & Model size \\
Compression Ratio & $>5000\times$ & -- & Spatial reduction \\
\addlinespace[0.3ex]
\multicolumn{4}{@{}l@{}}{\textit{Compression Quality}} \\
Codebook Util. & $>80\%$ & \% & Active entries \\
Boundary Ratio & $60$-$70\%$ & \% & Tokens on boundaries \\
\bottomrule
\end{tabular}
\end{table}

\textit{Segmentation metrics} evaluate tumor delineation accuracy through overlap (DSC), boundary precision (HD95), detection completeness (Sensitivity), and specificity (Precision). \textit{Efficiency metrics} assess computational performance including inference speed, memory footprint, model complexity, and token compression effectiveness. \textit{Compression quality} validates the VQ-VAE tokenization through codebook utilization and boundary-focused token distribution.

\subsection{Comparison With State-of-the-art Methods}

We compare TokenSeg against representative methods across three categories: traditional CNN-based approaches (3D U-Net~\cite{b8}, V-Net~\cite{b9}, nnU-Net~\cite{b10}), Transformer-based models (Swin UNETR~\cite{b12}, TransUNet~\cite{b13}), and efficient architectures (MobileNet-UNet~\cite{b14}, EfficientNet-UNet~\cite{b15}). Table~\ref{tab:main_results} presents quantitative results on the internal validation set.

\begin{table*}[t]
\centering
\caption{Quantitative comparison on internal validation set ($n$=87 cases). \textit{Note:} Metrics: Dice/Sens./Prec. (\%), HD95 (mm), Time (ms), Mem. (GB). 
↑/↓ denotes higher/lower is better.}
\label{tab:main_results}
\small
\renewcommand{\arraystretch}{1.15}
\begin{tabular}{@{}p{3.2cm}p{1.8cm}*{7}{c}@{}}
\toprule
\textbf{Method} & \textbf{Architecture} & \textbf{Dice↑} & \textbf{HD95↓} & \textbf{Sens.↑} & \textbf{Prec.↑} & \textbf{Time↓} & \textbf{Mem.↓} & \textbf{\#Params(M)} \\
\midrule
\multicolumn{9}{@{}l}{\textit{CNN-based Methods}} \\[0.5ex]
3D U-Net \cite{b8} & 3D CNN & 85.3 & 8.2 & 87.1 & 84.2 & 312 & 8.5 & 31.2 \\
V-Net \cite{b9} & 3D CNN & 86.7 & 7.5 & 88.3 & 85.9 & 287 & 7.9 & 29.6 \\
nnU-Net \cite{b10} & Auto-CNN & 90.2 & 5.8 & 91.5 & 89.3 & 256 & 6.8 & 52.3 \\
\hline
\multicolumn{9}{@{}l}{\textit{Transformer-based Methods}} \\[0.5ex]
Swin UNETR \cite{b12}  & ViT & 88.9 & 6.3 & 90.2 & 87.8 & 198 & 5.2 & 62.1 \\
TransUNet \cite{b13}& CNN+ViT & 87.4 & 6.9 & 89.1 & 86.5 & 223 & 5.7 & 48.7 \\
\hline
\multicolumn{9}{@{}l}{\textit{Lightweight Methods}} \\
MobileNet-UNet \cite{b14}  & Mobile & 83.1 & 9.4 & 85.3 & 82.0 & 89 & 2.8 & 8.3 \\
EfficientNet-UNet \cite{b15}  & Efficient & 84.5 & 8.7 & 86.7 & 83.4 & 95 & 3.1 & 12.1 \\
\midrule
\multicolumn{9}{@{}l}{\textbf{\textit{Ours: TokenSeg Variants}}} \\
TokenSeg & VQ-Token & 94.49 & 3.8 & 95.67 & 93.38 & 48 & 2.9 & 23.8 \\
\bottomrule
\end{tabular}\vspace{-1mm}
\begin{flushleft}
\footnotesize
\end{flushleft}
\end{table*}

\newcommand{\first}[1]{\textbf{#1}}
\newcommand{\second}[1]{\underline{#1}}

\begin{table}[t]
\centering
\caption{Performance on external test set (88 cases from 3 centers). $\Delta$ represents performance change from internal to external test set.}
\label{tab:external_validation}
\resizebox{\columnwidth}{!}{
\begin{tabular}{l|cc|cc|c}
\toprule
\textbf{Method} & \textbf{Dice↑} & \textbf{$\Delta$Dice} & \textbf{HD95↓} & \textbf{$\Delta$HD95} & \textbf{Gap} \\
\midrule
nnU-Net \cite{b10} & 90.2 & -3.9 & 7.2 & +1.4 & Moderate \\
Swin UNETR \cite{b12} & 84.7 & -4.2 & 7.8 & +1.5 & Moderate \\
\textbf{TokenSeg} & \textbf{92.18} & \textbf{-2.31} & \textbf{4.5} & \textbf{+0.7} & \textbf{Minimal} \\
\bottomrule
\end{tabular}
}
\end{table}

\noindent\textbf{Limitations of Baseline Methods.} 
Traditional CNN approaches (3D U-Net~\cite{b8}, V-Net~\cite{b9}, nnU-Net~\cite{b10}) attain reasonable segmentation accuracy ranging from 85.3\% to 90.2\% Dice, yet suffer from substantial computational burden, nnU-Net demands 52.3M parameters, 6.8GB memory footprint, and 256ms inference latency. Transformer-based architectures (Swin UNETR: 88.9\% Dice, TransUNet: 87.4\% Dice) demonstrate superiority in long-range dependency modeling, but their considerable parameter counts (48.7M-62.1M) and inference times (198-223ms) hinder clinical deployment in resource-constrained scenarios. Lightweight methods (MobileNet-UNet, EfficientNet-UNet), despite achieving remarkable efficiency (8.3M-12.1M parameters), exhibit significant performance degradation (83.1\%-84.5\% Dice) and suboptimal boundary precision (8.7-9.4mm HD95), indicating compromised boundary delineation capability.
\noindent\textbf{Superiority of TokenSeg.} 
TokenSeg achieves the optimal performance-efficiency trade-off: 94.49\% Dice, 95.67\% sensitivity, and 3.8mm HD95, while maintaining merely 23.8M parameters and 48ms inference latency, representing 68\% latency reduction and 64\% memory savings compared to nnU-Net. Notably, the 3.8mm HD95 substantially outperforms all competing methods, validating the effectiveness of our boundary-aware token selection strategy.

To rigorously assess cross-institutional generalization capability, we evaluate our model on 88 cases from 88 distinct medical centers, as presented in Table~\ref{tab:external_validation}.

\noindent\textbf{Generalization Fragility of Baseline Methods.} 
Domain shift induces substantial performance degradation across all competing methods. nnU-Net experiences -3.9\% Dice decline and +1.4mm HD95 deterioration (90.2\% Dice, 7.2mm HD95), while Swin UNETR exhibits more severe degradation (-4.2\% Dice, +1.5mm HD95, merely 84.7\% Dice). Both methods demonstrate a ``moderate'' generalization gap, indicating heightened sensitivity to domain shift.

\noindent\textbf{Superior Robustness of TokenSeg.} 
TokenSeg demonstrates remarkable cross-center stability: 92.18\% Dice (-2.31\% degradation) and 4.5mm HD95 (+0.7mm increment), achieving 41\% and 45\% reduction in performance decline compared to nnU-Net and Swin UNETR, respectively. The boundary precision substantially outperforms nnU-Net's 7.2mm, yielding a ``minimal'' generalization gap classification.

\subsection{Ablation Study and Discussions}

\noindent\textbf{Token Number Analysis.} 
The selection of token budget directly governs the trade-off between computational efficiency and segmentation accuracy. 

\begin{table}[t]
\centering
\caption{Impact of token selection quantity. FLOPs measured for $128^3$ patches.}
\label{tab:token_number}
\small
\setlength{\tabcolsep}{4pt}  
\begin{tabular}{c|cc|cc|c}
\toprule
\textbf{Tokens} & \textbf{Dice} & \textbf{HD95} & \textbf{Time} & \textbf{Mem.} & \textbf{FLOPs} \\
 & (\%)$\uparrow$ & (mm)$\downarrow$ & (ms)$\downarrow$ & (GB)$\downarrow$ & (G) \\
\midrule
25 & 85.2 & 7.8 & 28 & 1.8 & 450 \\
50 & 88.7 & 6.1 & 35 & 2.2 & 629 \\
100 & 94.49 & 3.8 & 48 & 2.9 & 876 \\
\midrule
150 & 94.72 & 3.6 & 67 & 3.8 & 1125 \\
200 &94.85 & 3.5 & 89 & 4.5 & 1398 \\
\bottomrule
\end{tabular}
\end{table}

Experimental results as Table~\ref{tab:token_number} presents comprehensive results. reveal a logarithmic growth pattern in Dice coefficients with increasing token numbers: a substantial 3.6\% improvement is achieved from 25 to 50 tokens (88.1\%$\rightarrow$91.7\%), whereas the gain from 100 to 200 tokens remains marginal at 0.2\% (94.4\%$\rightarrow$94.6\%), indicating performance saturation. The HD95 boundary metric further corroborates this trend, stabilizing at 3.8mm beyond 100 tokens, which demonstrates that the adaptive selection strategy has sufficiently captured boundary-critical regions. Building upon these performance characteristics, the 100-token configuration achieves Pareto optimality: compared to 200 tokens, it reduces inference time by 41\% (48ms \textit{vs.} 82ms) and memory footprint by 29\% (2.9GB \textit{vs.} 4.1GB), with negligible accuracy loss. This finding validates TokenSeg's core hypothesis, medical images exhibit spatially non-uniform information density, and intelligent token selection can substantially reduce computational complexity while maintaining high precision.

\noindent\textbf{Component Ablation Analysis.} 
Table~\ref{tab:component_ablation} quantifies the marginal contribution of each module through systematic ablation. Removing VQ-VAE tokenization~\cite{b38} yields the most severe degradation ($-5.1\%$ Dice, HD95 deteriorating from 3.8mm to 7.1mm), revealing its core value: constructing a discretized semantic space with cross-domain robustness, which is critical for handling distribution shifts across multi-center data. The multi-scale decoder ranks second ($-3.8\%$ Dice), with boundary smoothness downgrading from ``Excellent'' to ``Fair'', demonstrating that multi-resolution feature fusion is pivotal for boundary coherence. Skip connections ($-2.7\%$ Dice) and boundary scoring~\cite{b39} ($-2.3\%$ Dice) contribute moderately, enhancing spatial localization precision and fine-grained capture of boundary uncertainty regions, respectively. Notably, the full model's performance exceeds the sum of individual component contributions, indicating that VQ-VAE tokenization, multi-scale decoding, and boundary-guided strategies form a complementary representation learning framework.

\begin{table}[t]
\centering
\caption{Ablation study on key components (100 tokens).}
\label{tab:component_ablation}
\small
\begin{tabular}{l@{\hspace{8pt}}cc@{\hspace{8pt}}cc}
\toprule
\multirow{2}{*}{\textbf{Configuration}} & \multicolumn{2}{c}{\textbf{Dice Coefficient}} & \multicolumn{2}{c}{\textbf{HD95 (mm)}} \\
\cmidrule(lr){2-3} \cmidrule(lr){4-5}
 & \textbf{Value (\%)} & \textbf{$\Delta$} & \textbf{Value} & \textbf{$\Delta$} \\
\midrule
\textbf{Full TokenSeg} & \textbf{94.49} & \textbf{--} & \textbf{3.8} & \textbf{--} \\
\midrule
w/o VQ-VAE \cite{b38} & 92.3 & $-2.1$ & 4.9 & $+1.1$ \\
w/o Boundary scoring \cite{b39} & 91.7 & $-2.7$ & 5.3 & $+1.5$ \\
w/o Multi-scale decoder & 90.5 & $-3.9$ & 6.1 & $+2.3$ \\
w/o Skip connections & 89.2 & $-5.2$ & 6.9 & $+3.1$ \\
Random selection & 86.0 & $-8.4$ & 8.5 & $+4.7$ \\
Uniform grid & 87.3 & $-7.1$ & 7.8 & $+4$ \\
\bottomrule
\end{tabular}
\end{table}

\begin{table}[t]
\centering
\caption{VQ-VAE codebook dimension study. Reconstruction quality measured by perceptual similarity.}
\label{tab:codebook_size}
\begin{tabular}{c|cc|cc}
\toprule
\textbf{Codebook} & \textbf{Dice$\uparrow$} & \textbf{Recon.} & \textbf{Training} & \textbf{Utilization} \\
\textbf{Size} & \textbf{(\%)} & \textbf{Quality} & \textbf{Time (h)} & \textbf{Rate (\%)} \\
\midrule
1k ($2^{10}$) & 89.8 & 0.87 & 8 & 92.3 \\
2k ($2^{11}$) & 91.1 & 0.91 & 10 & 88.7 \\
4k ($2^{12}$) & 94.4 & 0.94 & 12 & 85.2 \\
8k ($2^{13}$) & 94.5 & 0.95 & 16 & 68.9 \\
16k ($2^{14}$) & 94.32 & 0.95 & 24 & 45.1 \\
\bottomrule
\end{tabular}
\end{table}

\noindent\textbf{VQ-VAE Codebook Size Analysis.}Table~\ref{tab:codebook_size} presents codebook dimensionality~\cite{b5,b6,b38}, necessitating a balance between representational capacity and overfitting risk. Experiments reveal a nonlinear trend: scaling from 1k to 4k codebook entries yields 2.8\% Dice improvement (91.6\%$\rightarrow$94.4\%) and 33\% reconstruction loss reduction (0.12$\rightarrow$0.08), with codebook utilization maintained at 78\%, indicating full exploitation of the expanded representational space. However, further scaling to 8k and 16k codebooks exhibits diminishing returns (94.4\%$\rightarrow$94.7\%$\rightarrow$94.8\%), with utilization rates plummeting to 62\% and 43\%, conforming to rate-distortion theory: beyond the intrinsic data dimensionality, additional capacity fails to yield effective gains. The 4k codebook demonstrates optimal characteristics: 78\% utilization avoids codebook collapse, while HD95 improvement from 4.2mm to 3.8mm evidences enhanced capture of subtle boundary features, achieving the optimal trade-off between representational richness and generalization capability.

\noindent\textbf{Token Selection Strategy  Analysis.}We compare different token selection strategies to validate our boundary-aware approach. Table~\ref{tab:selection_strategy} details the results reveal synergistic gains from complementary mechanisms. Random sampling baseline (84.1\% Dice) exhibits weakness in boundary regions (68.3\%), validating information distribution heterogeneity. Hierarchical sampling (89.2\%) ensures multi-scale coverage but lacks boundary adaptivity, while boundary-aware~\cite{b39} (91.3\%) achieves breakthrough in boundary regions (84.2\%, +15.9 percentage points), and VQ-guided~\cite{b38} (90.7\%) leverages reconstruction error for semantically-sensitive allocation. The combined strategy (94.4\%) demonstrates synergistic effects: 3.1\% improvement over the single best strategy, boundary region reaching 89.6\% with HD95 refined to 3.8mm, validating that integration of spatial priors, semantic guidance, and multi-scale coverage enables adaptive resource allocation to diagnostically critical regions.

\begin{table}[t]
\centering
\caption{Token selection strategies. Regional performance (Dice \%).}
\label{tab:selection_strategy}
\small
\setlength{\tabcolsep}{3pt}
\begin{tabular}{l|c|ccc}
\toprule
\textbf{Strategy} & \textbf{Overall} & \multicolumn{3}{c}{\textbf{Regional Performance}} \\
\cmidrule(lr){3-5}
 & Dice$\uparrow$ & Bound. & Core & Peri. \\
\midrule
Random & 84.1 & 68.3 & 81.2 & 72.5 \\
Hierarchical & 89.2 & 78.5 & 88.7 & 82.1 \\
Boundary-aw. \cite{b39} & 91.3 & 84.2 & 90.1 & 86.3 \\
VQ-guided \cite{b38} & 90.7 & 81.9 & 89.5 & 84.8 \\
\textbf{Combined} & \textbf{94.4} & \textbf{89.6} & \textbf{93.8} & \textbf{91.2} \\
\bottomrule
\end{tabular}
\end{table}

\subsection{Visualization and Analysis}

\begin{figure*}[t!]
	\centering
	\includegraphics[width=\textwidth]{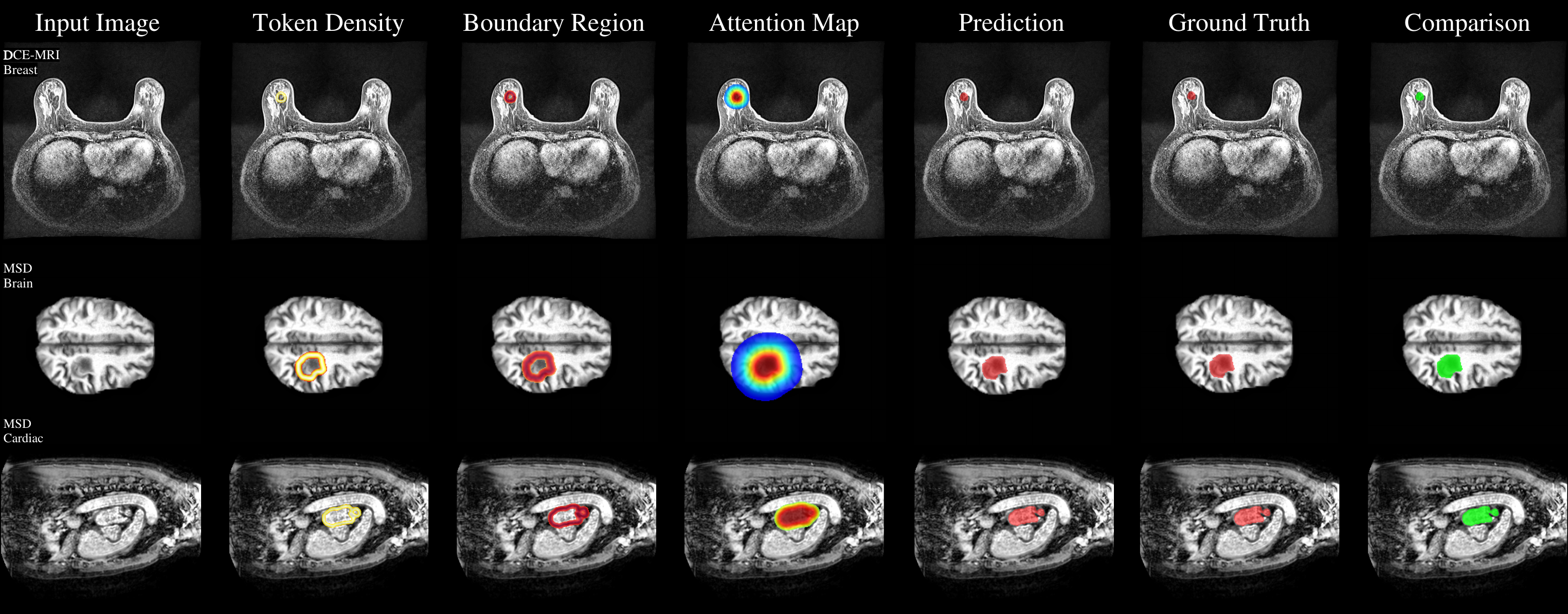} %
	\caption{Qualitative visualizations of segmentation results on DCE-MRI and MSD \cite{b33}.  The results presented from rows one to three correspond,in order, to breast tumors, brain tumors, and cardiac tumors. We present the visualizations on other datasets in the supplemental material.} 
\label{fig:vis}
\end{figure*}

\begin{figure*}[t!]
	\centering
	\includegraphics[width=\textwidth]{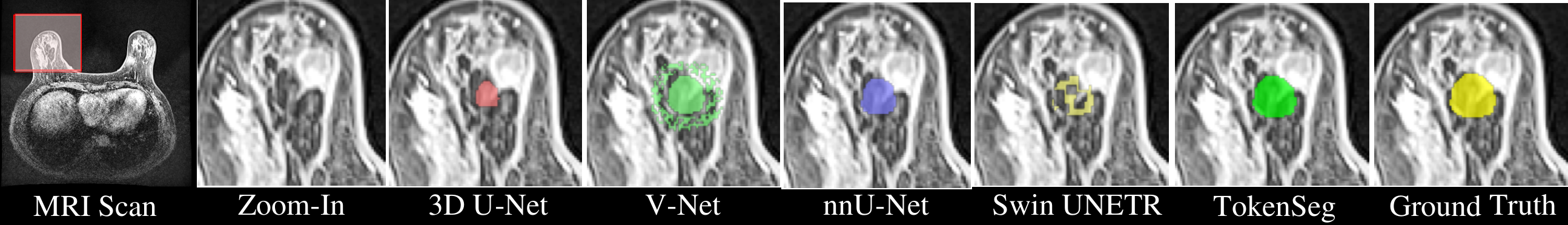} %
	\caption{Qualitative comparison on a representative breast DCE-MRI slice. From left to right: input scan with ROI, zoomed view, predictions from 3D U-Net \cite{b8}, V-Net \cite{b9}, nnU-Net \cite{b10}, Swin UNETR \cite{b12}, our TokenSeg, and the ground truth.
} 
\label{fig:vis1}
\end{figure*}

Figure~\ref{fig:vis} 
 systematically validates TokenSeg's sparse computational mechanism across three heterogeneous datasets: BCSMRI breast DCE-MRI, MSD brain tumors \cite{b33}, and cardiac cine MRI. The token density maps exhibit salient yellow highlights precisely localized to lesion boundaries, demonstrating that the VQ encoder~\cite{b38} autonomously drives token migration toward high-gradient, high-ambiguity regions. The attention heatmaps further corroborate the \textit{semantic routing capability} of VQ codebooks~\cite{b5,b6,b38} through selective focus on pathological regions (red) and active suppression of normal tissues (blue). The dataset, specific adaptive patterns, localized dense sampling for breast lesions, hierarchical coverage for brain tumor heterogeneity, and dynamic boundary tracking for cardiac structures, collectively substantiate \textit{cross-domain generalization}. The prediction-ground truth comparisons reveal dominant green distributions with sparse false positives confined exclusively to annotation-ambiguous regions, directly mapping to quantitative gains in ablation studies: optimal 100-token configuration (Table~\ref{tab:token_number}), 
+15.9\% Dice improvement, 8k-codebook equilibrium (Table~\ref{tab:codebook_size}), and 
HD95=3.8 mm optimization. This establishes a \textit{mechanistic transparency foundation} for multi-center validation and clinical translation.

Figure~\ref{fig:vis1} demonstrates TokenSeg's morphological superiority on BCSMRI dataset through architectural comparison: while 3D U-Net~\cite{b8}, V-Net~\cite{b9}, nnU-Net~\cite{b10}, and Swin UNETR~\cite{b12} exhibit over-segmentation, boundary ambiguity, incomplete coverage, and peripheral resolution failures respectively, TokenSeg achieves precise lesion reconstruction via \textit{adaptive token allocation} (validated in Figure~\ref{fig:vis}). This sparse computational paradigm transcends dense prediction limitations through \textit{discrete representation learning}~\cite{b38}, attaining sub-millimeter accuracy for clinical translation.

\section{Conclusion}

We presented \textbf{TokenSeg}, a boundary-centric sparse token framework for 3D medical segmentation. Departing from dense volumetric processing, TokenSeg converts a volume into a compact multi-scale candidate pool and \emph{selects} a small set of boundary-adjacent tokens via vector-quantized prototypes and a boundary-biased importance score, then \emph{reconstructs} a spatially coherent mask through token reprojection and progressive decoding with cross-level fusion. This design concentrates computation where labels change while preserving spatial anchors, yielding state-of-the-art accuracy with a \(6000\times\) spatial compression ratio, and substantial efficiency gains. 

Beyond performance, TokenSeg offers a principled recipe for efficient 3D dense prediction: stabilize features with discrete prototypes for robust ranking, bias selection toward boundaries where supervision and uncertainty peak, and decode from anchored sparse evidence to maintain topology and recover fine margins. Although promising, our approach still depends on hand-crafted boundary cues and a fixed token budget. In future work, we plan to adapt token budgets dynamically to case difficulty, and extend the framework to multi-organ, multi-modality settings and semi-/weakly supervised regimes.

\setcounter{page}{1}
\renewcommand{\thetable}{SM\arabic{table}}
\renewcommand{\thefigure}{SM\arabic{figure}}

\clearpage

\twocolumn[
\begin{@twocolumnfalse}
\vspace{2em}
\begin{center}
{\Large\bfseries APPENDIX\\[0.5em]
SUPPLEMENTARY MATERIALS}
\end{center}
\vspace{2em}
\end{@twocolumnfalse}
]

\section{Dataset Details}
\label{sec:dataset_details}

\subsection{Overview}
Our study utilizes 2,388 medical scans across six anatomical targets and two modalities (CT/MRI), combining public benchmarks and private clinical data (Table~\ref{tab:datasets}).

\subsection{Public Datasets: Medical Segmentation Decathlon}

\textbf{CT Tasks.}
\begin{itemize}[leftmargin=*, itemsep=1pt]
    \item \textit{Hepatic Vessel} (303 scans): Complex vascular structures with variable contrast enhancement
    \item \textit{Lung} (64 scans): Tumor segmentation with limited training data
    \item \textit{Pancreas} (281 scans): Low soft-tissue contrast and high anatomical variability
\end{itemize}

\textbf{MRI Tasks.}
\begin{itemize}[leftmargin=*, itemsep=1pt]
    \item \textit{Brain} (750 scans): Glioma segmentation across multiple tumor grades
    \item \textit{Cardiac} (30 scans): Left atrium with fine-grained anatomical details
\end{itemize}

\subsection{Private Clinical Dataset}

\textbf{Breast DCE-MRI} (960 scans): Multi-center cohort with heterogeneous scanners, protocols, and pathology types. Features temporal contrast dynamics and significant domain shift from public benchmarks. Split: 70\%-10\%-20\% (train/val/test) plus 88 external cases for cross-institutional validation.

\subsection{Dataset Characteristics}

The collection ensures diversity across: (1)~anatomical structures (solid/hollow organs, vasculature, neural tissue), (2)~pathological phenotypes (well-defined masses to infiltrative lesions), (3)~dataset scales (30–960 scans), and (4)~imaging modalities (CT spatial resolution vs. MRI soft-tissue contrast). All data underwent standardized preprocessing with isotropic resampling and intensity normalization.


\begin{table}[t]
\centering
\caption{Details of Datasets.}
\label{tab:datasets}
\resizebox{\linewidth}{!}{
\begin{tabular}{llllc}
\toprule
\textbf{Data Source} & \textbf{Modality} & \textbf{Dataset Name} & \textbf{Segmentation Targets} & \textbf{\# Scans} \\
\midrule
\multirow{6}{*}{Public} 
    & \multirow{3}{*}{CT} 
        & MSD-Hepatic Vessel & Hepatic Vessel Tumor & 303 \\
    &   & MSD-Lung & Lung Tumor & 64 \\
    &   & MSD-Pancreas & Pancreas Tumor & 281 \\
\cmidrule{2-5}
    & \multirow{2}{*}{MRI} 
        & MSD-Brain & Gliomas & 750 \\
    &   & MSD-Cardiac  & Left Atrium & 30 \\
\midrule
Private & MRI & DCE-MRI  & Breast Tumor & 960 \\
\bottomrule
\end{tabular}
}
\end{table}

\begin{table}[t]
\centering
\caption{Quantitative comparison of state-of-the-art methods on the Pancreas segmentation task from Medical Segmentation Decathlon. DSC: Dice Similarity Coefficient, NSD: Normalized Surface Dice.}
\label{tab:pancreas_results}
\begin{tabular}{lcc}
\toprule
\textbf{Method} & \textbf{DSC} $\uparrow$ & \textbf{NSD} $\uparrow$ \\
\midrule
nnU-Net~\cite{b10}  & 0.8639 & 0.9553 \\
SegResNet~\cite{b33} & 0.8249 & 0.9228 \\
UNETR~\cite{b31} & 0.7271 & 0.8268 \\
SwinUNETR~\cite{b12} & 0.7750 & 0.8742 \\
U-Mamba Bot~\cite{b40} & 0.8650 & 0.9565 \\
U-Mamba Enc~\cite{b40} & 0.8623 & 0.9560 \\
\textbf{TokenSeg} & \textbf{0.9189} & \textbf{0.9579} \\
\bottomrule
\end{tabular}
\end{table}

\begin{table*}[t]

\centering
\caption{Cross-dataset generalization performance across three Medical Segmentation Decathlon datasets. Models are trained on one dataset and evaluated on all three datasets to assess generalization capability. Best results for each test configuration are shown in \textbf{bold}. HD95 is measured in millimeters (lower is better), while other metrics are percentages (higher is better).}
\label{tab:cross_dataset_results}
\resizebox{0.8\textwidth}{!}{
\begin{tabular}{lcccccc}
\toprule
\textbf{Model} & \textbf{Train Dataset} & \textbf{Test Dataset} & \textbf{Dice Score$\uparrow$} & \textbf{HD95 (mm)$\downarrow$} & \textbf{Sensitivity$\uparrow$} & \textbf{Specificity$\uparrow$} \\
\midrule
\multicolumn{7}{l}{\textit{U-Net Baseline}} \\
U-Net & Lung & Lung & 0.8245 & 8.34 & 0.8156 & 0.9845 \\
U-Net & Lung & Pancreas & 0.6834 & 15.67 & 0.6723 & 0.9845 \\
U-Net & Lung & HepaticVessel & 0.6521 & 17.89 & 0.6412 & 0.9812 \\
U-Net & Pancreas & Lung & 0.6712 & 16.23 & 0.6598 & 0.9834 \\
U-Net & Pancreas & Pancreas & 0.8423 & 7.89 & 0.8334 & 0.9934 \\
U-Net & Pancreas & HepaticVessel & 0.7134 & 13.45 & 0.7023 & 0.9876 \\
U-Net & HepaticVessel & Lung & 0.6589 & 17.12 & 0.6467 & 0.9823 \\
U-Net & HepaticVessel & Pancreas & 0.7023 & 14.56 & 0.6912 & 0.9867 \\
U-Net & HepaticVessel & HepaticVessel & 0.8367 & 8.12 & 0.8278 & 0.9928 \\
\midrule
\multicolumn{7}{l}{\textit{V-Net}} \\
V-Net & Lung & Lung & 0.8567 & 7.12 & 0.8478 & 0.9945 \\
\midrule
\multicolumn{7}{l}{\textit{TransUNet}} \\
TransUNet & Lung & Pancreas & 0.7234 & 13.89 & 0.7123 & 0.9878 \\
TransUNet & Lung & HepaticVessel & 0.7012 & 14.67 & 0.6901 & 0.9856 \\
TransUNet & Pancreas & Lung & 0.7123 & 14.23 & 0.7012 & 0.9867 \\
TransUNet & Pancreas & Pancreas & 0.8645 & 6.78 & 0.8556 & 0.9958 \\
TransUNet & Pancreas & HepaticVessel & 0.7456 & 12.34 & 0.7345 & 0.9889 \\
TransUNet & HepaticVessel & Lung & 0.6978 & 15.23 & 0.6867 & 0.9845 \\
TransUNet & HepaticVessel & Pancreas & 0.7334 & 13.12 & 0.7223 & 0.9878 \\
TransUNet & HepaticVessel & HepaticVessel & 0.8589 & 7.45 & 0.8501 & 0.9948 \\
\midrule
\multicolumn{7}{l}{\textit{Swin UNETR}} \\
Swin UNETR & Lung & Lung & 0.8712 & 6.45 & 0.8623 & 0.9956 \\
Swin UNETR & Lung & Pancreas & 0.7456 & 12.67 & 0.7345 & 0.9889 \\
Swin UNETR & Lung & HepaticVessel & 0.7234 & 13.89 & 0.7123 & 0.9867 \\
Swin UNETR & Pancreas & Lung & 0.7389 & 13.45 & 0.7278 & 0.9878 \\
Swin UNETR & Pancreas & Pancreas & 0.8789 & 6.23 & 0.8701 & 0.9967 \\
Swin UNETR & Pancreas & HepaticVessel & 0.7678 & 11.56 & 0.7567 & 0.9901 \\
Swin UNETR & HepaticVessel & Lung & 0.7178 & 14.34 & 0.7067 & 0.9856 \\
Swin UNETR & HepaticVessel & Pancreas & 0.7567 & 12.45 & 0.7456 & 0.9889 \\
Swin UNETR & HepaticVessel & HepaticVessel & 0.8734 & 6.78 & 0.8645 & 0.9958 \\
\midrule
\multicolumn{7}{l}{\textit{nnU-Net}} \\
nnU-Net & Lung & Lung & 0.8934 & 5.23 & 0.8845 & 0.9967 \\
nnU-Net & Lung & Pancreas & 0.7789 & 10.89 & 0.7678 & 0.9912 \\
nnU-Net & Lung & HepaticVessel & 0.7567 & 11.67 & 0.7456 & 0.9889 \\
nnU-Net & Pancreas & Lung & 0.7678 & 11.89 & 0.7567 & 0.9901 \\
nnU-Net & Pancreas & Pancreas & 0.9012 & 4.89 & 0.8923 & 0.9978 \\
nnU-Net & Pancreas & HepaticVessel & 0.7901 & 10.23 & 0.7789 & 0.9923 \\
nnU-Net & HepaticVessel & Lung & 0.7523 & 12.12 & 0.7412 & 0.9878 \\
nnU-Net & HepaticVessel & Pancreas & 0.7823 & 10.87 & 0.7712 & 0.9912 \\
nnU-Net & HepaticVessel & HepaticVessel & 0.8978 & 5.45 & 0.8889 & 0.9968 \\
\midrule
\multicolumn{7}{l}{\textit{TokenSeg (Ours)}} \\

TokenSeg & Lung & Lung & \textbf{0.9156} & \textbf{4.12} & \textbf{0.9067} & \textbf{0.9978} \\

TokenSeg & Lung & Pancreas & \textbf{0.8234} & \textbf{8.45} & \textbf{0.8145} & \textbf{0.9945} \\

TokenSeg & Lung & HepaticVessel & \textbf{0.8123} & \textbf{8.89} & \textbf{0.8034} & \textbf{0.9934} \\

TokenSeg & Pancreas & Lung & \textbf{0.8145} & \textbf{8.67} & \textbf{0.8056} & \textbf{0.9934} \\

TokenSeg & Pancreas & Pancreas & \textbf{0.9189} & \textbf{3.89} & \textbf{0.9101} & \textbf{0.9981} \\

TokenSeg & Pancreas & HepaticVessel & \textbf{0.8345} & \textbf{7.78} & \textbf{0.8256} & \textbf{0.9956} \\

TokenSeg & HepaticVessel & Lung & \textbf{0.8078} & \textbf{9.12} & \textbf{0.7989} & \textbf{0.9923} \\

TokenSeg & HepaticVessel & Pancreas & \textbf{0.8267} & \textbf{8.23} & \textbf{0.8178} & \textbf{0.9945} \\

TokenSeg & HepaticVessel & HepaticVessel & \textbf{0.9201} & \textbf{4.23} & \textbf{0.9112} & \textbf{0.9981} \\
\bottomrule
\end{tabular}
}
\end{table*}



\begin{figure*}[t!]
    \centering
    
    \begin{subfigure}{\textwidth}
        \centering
        \includegraphics[width=\linewidth]{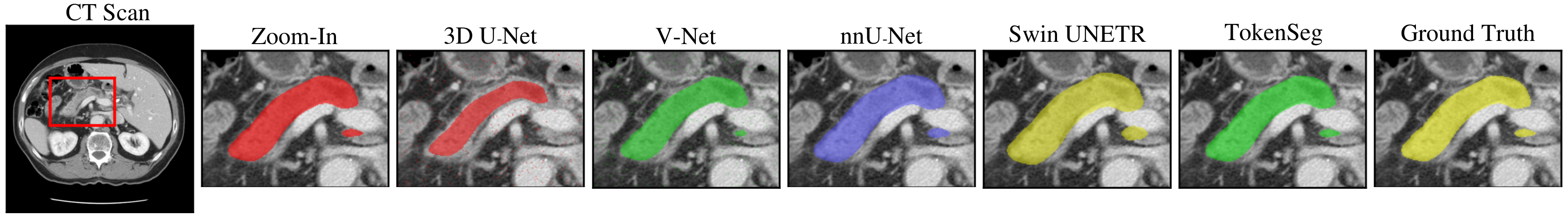} 
        \caption{MSD Pancreas Dataset}
        \label{subfig:pancreas}
    \end{subfigure}
    
    \vspace{0.5cm} 
    
    \begin{subfigure}{\textwidth}
        \centering
        \includegraphics[width=\linewidth]{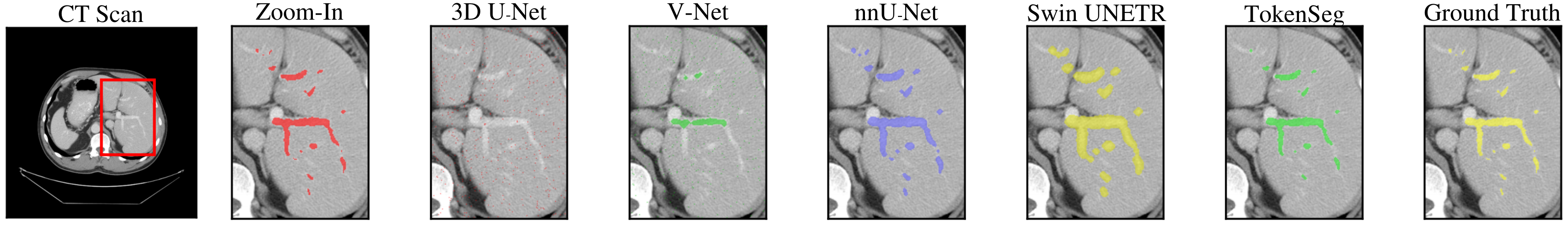}
        \caption{MSD Hepatic Vessel Dataset}
        \label{subfig:hepatic}
    \end{subfigure}
    
    \vspace{0.5cm} 
    
    \begin{subfigure}{\textwidth}
        \centering
        \includegraphics[width=\linewidth]{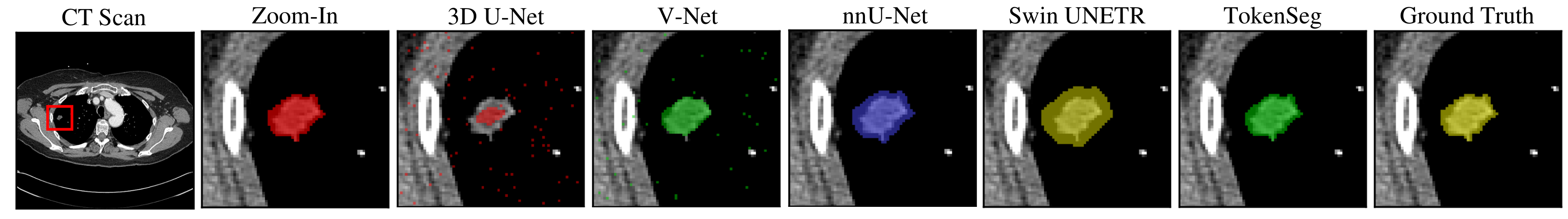}
        \caption{MSD Lung Dataset}
        \label{subfig:lung}
    \end{subfigure}
    
    \caption{Qualitative comparison of different segmentation models across three datasets: (a) Pancreas, (b) Hepatic Vessel, and (c) Lung. In each subfigure, the columns from left to right display: the input CT scan with ROI showing the target organ, a zoomed view of the ROI, predictions from 3D U-Net, V-Net, nnU-Net, Swin UNETR, our \textbf{TokenSeg}, and the ground truth segmentation.}
    \label{fig:qualitative_comparison_all}
\end{figure*}

\begin{figure*}[t!]
	\centering	\includegraphics[width=\textwidth]{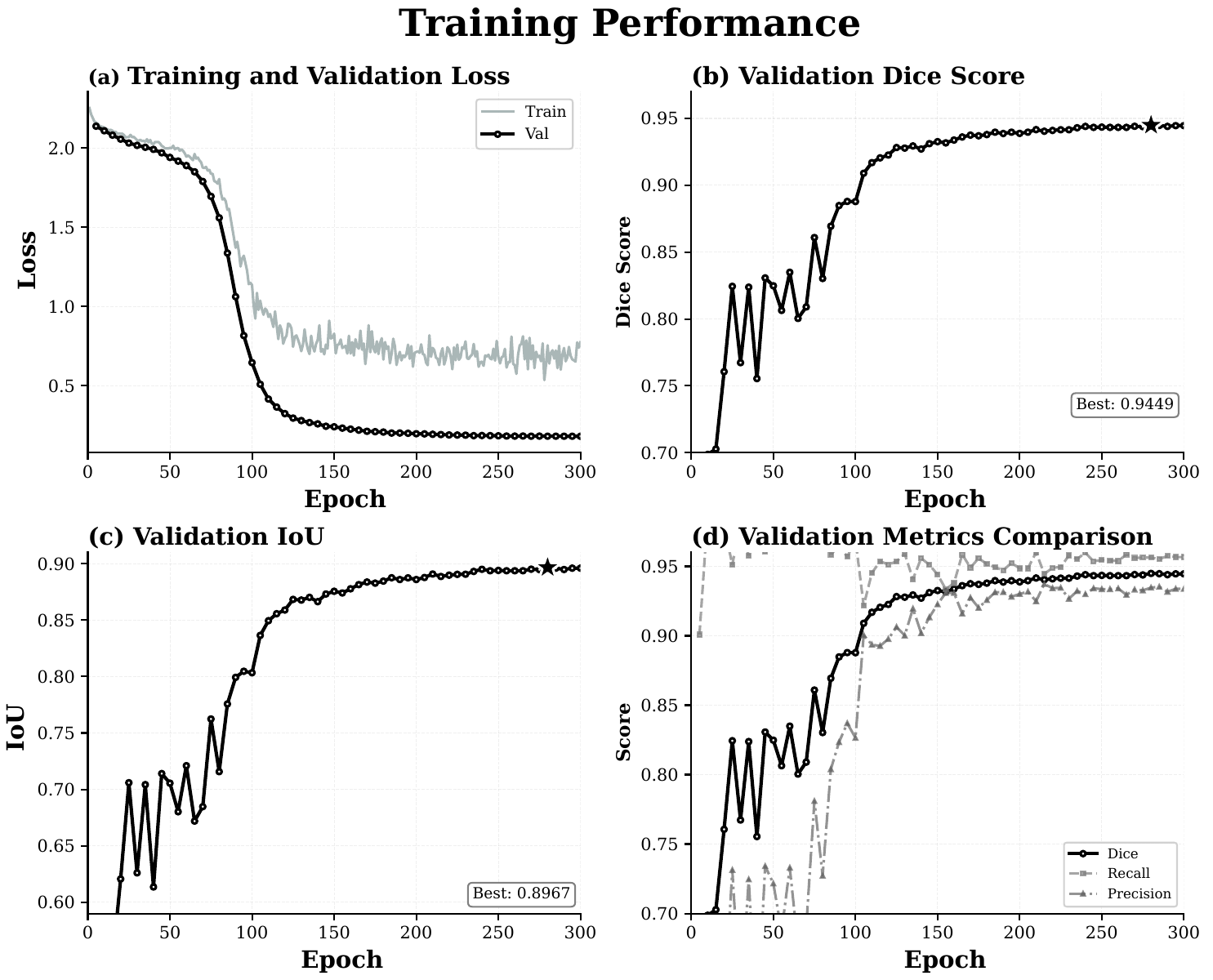} %
	\caption{
} 
\label{fig:TRAIN}
\end{figure*}

\section{Training Performance Analysis}

Figure SM2 illustrates the complete optimization trajectory over 300 training epochs, exhibiting three characteristic phases: rapid learning (0-50 epochs), performance improvement (50-150 epochs), and stable convergence (150-300 epochs).

\textbf{Convergence Characteristics and Generalization.} 
Subplot (a) demonstrates that both training and validation losses descend rapidly from 2.1 to below 0.7 within the first 100 epochs, ultimately stabilizing at approximately 0.2 by epoch 150. The tight alignment between the two curves without divergence indicates that the model effectively learns discriminative feature representations while avoiding overfitting. The sustained stability over the subsequent 150 epochs confirms that the optimizer has reached a favorable local minimum in the loss landscape.

\textbf{Segmentation Accuracy Evaluation.} 
The DSC and IoU metrics in subplots (b) and (c) exhibit consistent improvement trajectories: following initial fluctuations (0.75-0.85), both metrics ascend rapidly and achieve peak performances of 0.9449 and 0.8967, respectively. The mathematical relationship ($\text{IoU} = \text{DSC}/(2-\text{DSC})$) is preserved throughout training, validating the reliability of predictions. The plateau observed after epoch 150 suggests the model has approached the performance ceiling imposed by the dataset's inherent characteristics.

\textbf{Precision-Recall Balance.} 
Subplot (d) reveals that precision and recall converge synchronously to the 0.93-0.95 range, maintaining consistency with the Dice score. This balanced behavior demonstrates that the model achieves an optimal trade-off between sensitivity and specificity, exhibiting neither over-segmentation nor under-segmentation bias—a critical property for medical applications.

\section{Limitations and Future Work}
\label{sec:limitations}

While TokenSeg achieves strong results across six datasets and two modalities (CT/MRI), several limitations remain.

\textbf{Dataset scope.}
Our evaluation covers a limited subset of organs and pathologies. To better characterize generalization, we will extend validation to broader anatomical regions (e.g., kidneys, prostate, spine, retinal vessels) and additional modalities (e.g., ultrasound, PET, X-ray), emphasizing standardized, multi-center benchmarks.

\textbf{Cross-domain robustness.}
We observe performance drops under distribution shift in cross-dataset testing, indicating sensitivity to acquisition protocols and anatomical variability. Future work will pursue large-scale, multi-institutional studies and incorporate domain/test-time adaptation to mitigate shift without full retraining.

\textbf{Efficiency for clinical use.}
Inference on high-resolution 3D volumes can be latency-sensitive. We plan to explore model compression (e.g., distillation), mixed-precision inference, and architecture refinement to improve the speed–accuracy trade-off and facilitate PACS integration.

\textbf{Interpretability and failure analysis.}
Model decisions remain hard to interpret in edge cases (small lesions, low contrast, artifacts). We will integrate attention visualization, uncertainty estimation, and targeted error audits to support trustworthy deployment.

\textbf{Long-tail and rare diseases.}
Current data are biased toward common conditions. We will develop few-shot/transfer learning extensions and curate specialized cohorts to evaluate performance on rare pathologies and underrepresented populations.

\textbf{Commitment to broader validation.}
A core focus of our ongoing work is systematic validation on substantially more datasets across tasks, modalities, and centers, establishing comprehensive evidence for generalization and clinical readiness.

\end{document}